\documentclass{svproc}
\usepackage{url}
\usepackage{graphicx}%
\usepackage{multirow}%

\usepackage{amsthm}
\usepackage{amsmath,amssymb,amsfonts}%
\usepackage{mathrsfs}%
\usepackage[title]{appendix}%
\usepackage{xcolor}%
\usepackage{textcomp}%
\usepackage{manyfoot}%
\usepackage{booktabs}%
\usepackage{algorithmicx}%
\usepackage{algpseudocode}%
\usepackage{listings}
\usepackage{subfig}
\usepackage{tikz}
\usepackage{cleveref}
\usepackage{float}
\usetikzlibrary{arrows.meta}
\usetikzlibrary{calc}

\usepackage{algorithm,algpseudocode}

\usepackage{pgfplots}
\pgfplotsset{compat=1.16} % Use a supported compatibility version

\newcommand{\nullspace}{\text{null}}

\makeatletter
\def\UTFviii@defined#1{%
	\ifx#1\relax
	!!FIXME!!%
	\else
	\expandafter#1%
	\fi
}
\makeatother

\begin{document}
\mainmatter              % start of a contribution
\title{A Constrained Saddle Search Approach for Constructing Singular and Flexible Bar Frameworks}
\titlerunning{Constrained Saddle Search for Singular and Flexible Bar Frameworks}  % abbreviated title (for running head)
%                                     also used for the TOC unless
%                                     \toctitle is used
%
\author{Xuenan Li\inst{1} \and Mihnea Leonte\inst{1} \and Christian D. Santangelo\inst{2} \and Miranda Holmes-Cerfon\inst{3}}
\authorrunning{Xuenan Li et al.} % abbreviated author list (for running head)
%
%%%% list of authors for the TOC (use if author list has to be modified)
%\tocauthor{Ivar Ekeland, Roger Temam, Jeffrey Dean, David Grove,
%Craig Chambers, Kim B. Bruce, and Elisa Bertino}
%
\institute{Columbia University, New York, NY 10027, USA\\
\email{xl3383@columbia.edu}\and
Syracuse University, Syracuse, NY 13244, USA\and
The University of British Columbia, Vancouver, BC V6T 1Z4, Canada}

\maketitle              % typeset the title of the contribution

\begin{abstract}
	Singularity analysis is essential in robot kinematics, as singular configurations cause loss of control and kinematic indeterminacy. This paper models singularities in bar frameworks as saddle points on constrained manifolds. Given an under-constrained, non-singular bar framework, by allowing one edge to vary its length while fixing lengths of others, we define the squared length of the free edge as an energy functional and show that its local saddle points correspond to singular and flexible frameworks. Using our constrained saddle search approach, we identify previously unknown singular and flexible bar frameworks, providing new insights into singular robotics design and analysis.
\end{abstract}


\section{Introduction}\label{sec1:introduction}
\vspace{-2ex}
Singularity analysis has long been a fundamental topic in robot kinematics and remains an active area of research \cite{muller2015kinematic}. When a robotic system reaches a singular configuration, certain infinitesimal flexes fail to correspond to realizable deformations, leading to a loss of control over the system’s configuration space and resulting in kinematic indeterminacy. While singularities are often associated with challenges due to the loss of control, they can sometimes offer unique opportunities for exploration in robotic systems since a robot can transit between different kinematic configurations without requiring complex reconfiguration \cite{nakamura1986inverse}.

\begin{figure}[!htb]
	\centering
	\includegraphics[width=0.845\linewidth]{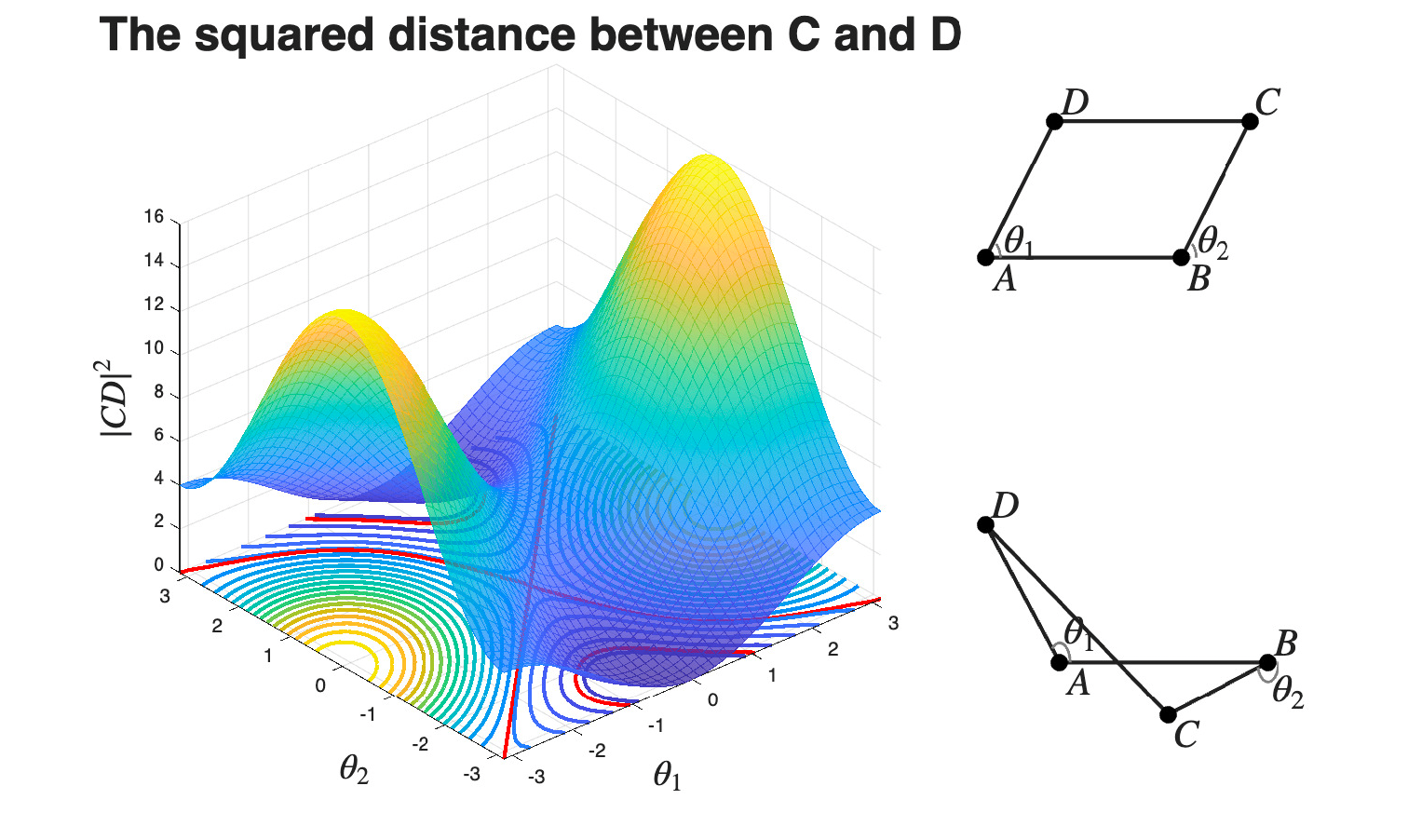}
	\caption{The saddle behavior in the four-bar linkage at $\theta_1 = \theta_2 = 0$.}
	\label{fig:saddle-branch}
\end{figure}

A common and systematic approach to studying singularities in robotic systems is through the analysis of the Jacobian matrix derived from geometric constraints. To illustrate this, we consider a well-known example: the four-bar linkage in \cref{fig:saddle-branch}, a simple yet widely studied singular and flexible bar framework (see e.g. \cite{mannattil2022thermal}). This particular 2D bar framework consists of four vertices and four edges, with two pairs of opposite edges having equal lengths. When the edges $AB$ and $AD$ do not align into a straight line (as illustrated in the right figures of \cref{fig:saddle-branch}), the four-bar linkage is non-singular and the non-trivial deformations that keep the lengths of all edges form a 1-dimensional smooth manifold. The only degree of freedom in this motion can be characterized as $\theta_1$ in \cref{fig:saddle-branch} ($\theta_2$ is determined by $\theta_1$). However, when the edges $AB$ and $AD$ align into a straight line with $\theta_1 = \theta_2=0$ (or $\pi$), the four-bar linkage becomes singular. The $4 \times 8$ Jacobian matrix of the bar framework's forward kinematics, also known as the rigidity matrix, loses rank. In fact, the Jacobian matrix is rank 3 and its null space has a two-dimensional subspace of infinitesimal flexes that are not infinitesimal rigid body motions. However, not all infinitesimal flexes in this subspace originate from nonlinear flexes that keep the lengths of all bars. The configuration space at this singular structure has two branches of 1-dimensional nonlinear flexes (these two flexes can be viewed as the two deformations in \cref{fig:saddle-branch} with small $\theta_1$). The singularity among four-bar linkages can be viewed as a branch point where some infinitesimal flexes cannot be realized by nonlinear flexes.

Let us analyze this singular bar framework from a different perspective by allowing the edge $CD$ to vary its length. Without the edge $CD$, we have an extra degree of freedom $\theta_2$ to deform the bar framework without changing the lengths of edges $AD, AB$ and $BC$. When viewing the squared length of edge $CD$ as a function of $\theta_1$ and $\theta_2$ in \cref{fig:saddle-branch}, we can see that there are two saddle points at $\theta_1 = \theta_2 = 0$ (or $\pi$). Both saddle points correspond to the singular structures where edges $AB$ and $AD$ are colinear. In fact, the two branches of nonlinear flexes correspond to the two branches in the level set at $\theta_1 = \theta_2 =0$. The singularity of the bar framework originates from the level set's singularity: though the 2-dimensional non-trivial infinitesimal flexes can be parameterized by $\theta_1$ and $\theta_2$, only the directions tangent to the level set along $\theta_1 = \pm \theta_2$ correspond to realizable nonlinear flexes within the constraint set. 
% Begin a figure environment

Now let us generalize such a perspective of viewing singularity as saddle points on a constrained manifold to general flexible bar frameworks. We start with an under-constrained bar framework in $\mathbb{R}^d$ with $n$ vertices and $m$ edges ($nd > m + d(d+1)/2$). We assume that this given bar framework is non-singular and the configuration space of deformations that do not change the lengths of bars forms a $(nd-m)$-dimensional manifold. To avoid rigid body motions, we choose a \textit{pinning scheme} so that the non-trivial nonlinear flexes form a $\left(nd-m-d(d+1)/2\right)$-dimensional manifold. Then, we allow a certain edge to vary its length and take the squared length of this free edge as the energy functional. By searching for index-$k$ saddle points of this energy functional on the constrained manifold where the lengths of the rest edges are fixed, we are guaranteed to find a singular and flexible bar framework. Without loss of generality, we choose the first edge as the free edge, and our constrained saddle search problem can be stated as follows:
\begin{align}
	\text{find index-}k \text{ saddles for }& f_1(\vec{p}) = |\vec{p}_{1,1}-\vec{p}_{1,2}|^2, \label{eqn:saddle-search}\\
	\text{s.t. }& f_i(\vec{p}) = |\vec{p}_{i,1}-\vec{p}_{i,2}|^2 - l_i^2 = 0,\quad i=2,\dots,m. \label{eqn:saddle-search-cst}\\
	& g_j(\vec{p}) = \vec{0},\quad j=1,\dots, d(d+1)/2,\label{eqn:pinning}
\end{align}
where $\vec{p}$ is a collection of vertices. The constraint functions $f_i(\vec{p})$ with $i=1,\dots,m$ are the squared length of the $i$th edge and $l_i$ is the length of the $i$th edge. We choose a pinning scheme $g_j(\vec{p})=\vec{0}$ with $j=1,\dots,d(d+1)/2$. In general, one can randomly pin $d(d+1)/2$ coordinates of $\vec{p}$. 

Our main result, \cref{thm:singular-flexible}, establishes that under certain mild conditions, a non-degenerate index-$k$ saddle point corresponds to a singular and flexible bar framework. We also provide a constrained index-$k$ saddle search algorithm and reveal new singular structures that do not have the symmetry of the four-bar linkage. We highlight that this approach offers significant flexibility in designing singular structures, as it does not require any inherent symmetry, unlike the commonly studied singular structures. It is worth mentioning that instead of searching for saddle points, if we optimize (minimize or maximize) $f_1(\vec{p})$ under the constraints in \cref{eqn:saddle-search-cst,eqn:pinning}, then we obtain rigid bar frameworks with second-order rigidity \cite{li2025constrained}. Although these bar frameworks are rigid, they can still be viewed as singular due to the existence of non-trivial first-order flexes that fail to integrate into nonlinear flexes.

\section{Preliminiaries and notation}\label{sec2:prelim}
\vspace{-2ex}

\textbf{Singular and flexible bar frameworks:} We start with a bar framework with $n$ vertices and $m$ edges: we denote the vertex set $\{\vec{p}_1, \vec{p}_2, \dots, \vec{p}_n\} \subseteq \mathbb{R}^d$ and the edge set $\{E_1,E_2,\dots,E_m\}$. A bar framework is \textit{flexible} if there are deformations other than rigid body motions that keep the lengths of all bars. For a flexible bar framework, we call the deformations that keep the lengths of all bars \textit{nonlinear flexes}, also known as \textit{mechanisms}. 

A simple counting argument, dating back to Maxwell \cite{maxwell1864calculation}, is commonly used to suggest whether a bar framework is flexible. Maxwell's counting argument states that for a flexible bar framework, the number of edges should not exceed the number of degrees of freedom minus the number of rigid body motions, i.e. a flexible bar framework in $\mathbb{R}^d$ with $n$ vertices and $m$ edges should satisfy $m + d(d+1)/2 < nd$. We categorize these bar frameworks \textit{under-constrained}. It is also tempting to consider there are $nd-m-d(d+1)/2$ internal degrees of freedom for nonlinear flexes among under-constrained bar frameworks. However, such counting arguments do not always work. In fact, it only works when there is no \textit{singularity} in the configuration space.

Let us mathematically explain the flexibility and singularity among under-constrained bar frameworks. A useful notation is to collect all vertices as a vector $\vec{p} = (\vec{p}_1,\dots,\vec{p}_n) \in \mathbb{R}^{nd}$ and also denote $\vec{p}_{i,1},\vec{p}_{i,2}$ as the two ends of the $i$th edge $E_i$. Using this notation, we can view a nonlinear flex of a bar framework as a smooth deformation $\vec{p}(t): I \to \mathbb{R}^{nd}$ defined on a small interval $I$ around 0, such that $ \vec{p}(0) = \vec{p}$ at the given bar framework and the lengths of the edges remain unchanged on $I$, i.e. the deformation satisfies that for all $i=1,\dots,m$
\begin{align}
	f_i(\vec{p}(t)) &= 0, \quad \text{for all } t \in I, \quad \text{with }f_i(\vec{p}) := |\vec{p}_{i,1} - \vec{p}_{i,2}|^2 - l_i^2, \label{eqn:nonlinear-flex}
\end{align}
where $l_i$ represents the length of the $i$th edge. Therefore, a nonlinear flex is a smooth curve on the level set
\begin{align}
	\mathcal{M}(\vec{p})=\{\vec{q}\in\mathbb{R}^{nd} \:|\: f_i(\vec{q}) = 0,i = 1,\dots,m\}.\label{eqn:level-set}
\end{align}
A bar framework is \textit{flexible} if the level set $\mathcal{M}(\vec{p})$ includes non-trivial deformations other than rigid body motions. A flexible bar framework is \textit{singular} if the level set $\mathcal{M}(\vec{p})$ is not a $(nd-m)$-dimensional smooth manifold. A necessary condition for the singularity in bar frameworks is that $\nabla f_i(\vec{p})$ with $i=1,\dots,m$ are linearly dependent. In fact, we call the following Jacobian matrix
\begin{align}
	\vec{R}(\vec{p}) = \begin{pmatrix}                                                                                                                                                                                                                                                                                                                                                                                                                                                                                                                                                                                                                                                                                                                                                                                                                                                                                                                                          
		\nabla^T f_1(\vec{p}) & \dots&                                                                                                                                                                                                                                                                                                                                                                                                                                                                                                                                                                                                                                                                                                                                                                                                                                                                                                                                             
		\nabla^T f_m(\vec{p})                                                                                                                                                                                                                                                                                                                                                                                                                                                                                                                                                                                                                                                                                                                                                                                                                                                                                                                                                      
	\end{pmatrix}^T \in \mathbb{R}^{m\times nd}\label{eqn:rigidity-matrix}                         
\end{align}                                                                       rigidity matrix, where $\nabla^T$ represents the gradient as a column vector. Therefore, a flexible bar framework is singular if and only if $\vec{R}(\vec{p})$ loses rank and some null vectors of $\vec{R}(\vec{p})$ cannot be realized by smooth curves in $\mathcal{M}(p)$.

Let us also introduce the non-trivial flexes. For a nonlinear flex $\vec{p}(t)$, its velocity $\vec{v}=\vec{p}'(0)\in \mathbb{R}^{nd}$ satisfies the linear relation $(\vec{p}_{i,1}- \vec{p}_{i,2}) \cdot (\vec{v}_{i,1}- \vec{p}_{i,2}) = 0$ with $i = 1,2,\dots,m$, meaning that $\vec{v} \in \nullspace \: \vec{R}(\vec{p})$. A vector $\vec{v} \in \mathbb{R}^{nd}$ in the null space of the rigidity matrix, i.e. $\vec{v} \in \nullspace \: \vec{R}(\vec{p})$ is an \textit{infinitesimal flex}, also known as first-order flexes. We notice that rigid body motions are trivial nonlinear flexes, and their derivatives are trivial infinitesimal flexes. For a $d$-dimensional bar framework, rigid body motions form a smooth manifold with dimension $d(d+1)/2$. The tangent space of this manifold contains all infinitesimal rigid body motions and we denote it as $\mathcal{T}(\vec{p})$. We call a nonlinear flex non-trivial if its velocity satisfies $\vec{v} \notin \mathcal{T}(\vec{p})$. It is worth noting that for a singular and flexible bar framework, a non-trivial infinitesimal flex may not come from a nonlinear flex. A necessary condition for a non-trivial infinitesimal flex to come from a nonlinear flex is known as the second-order stress test (see e.g. \cite{connelly1996second}). In this paper, we are particularly interested in these singular and flexible bar frameworks.

\textbf{Index-$k$ saddle points on constrained manifolds:} We introduce saddle points for an energy functional on a constrained manifold (see e.g. \cite{yin2022constrained}). Given a smooth energy functional $E(\vec{x}):\mathbb{R}^d \rightarrow \mathbb{R}$ and smooth constraint functions $\vec{c}(\vec{x}):\mathbb{R}^d \rightarrow \mathbb{R}^m$, we consider the energy functional $E(\vec{x})$ subject to the equality constraints $\vec{c}(\vec{x})=\vec{0}$. To avoid technicality difficulty, we always assume that
\begin{align}
	\nabla c_1(\vec{x}), \dots,\nabla c_m(\vec{x}) \text{ are linearly independent}.\label{eqn:LICQ}
\end{align}
Therefore, the level set $\vec{c}(\vec{x})=\vec{0}$ forms a $(d-m)$-dimensional smooth manifold. This linear independence condition is also known as the \textit{linear independence constraint qualification} (LICQ) condition in constrained optimization \cite{nocedal1999numerical}. The normal space of the level set $\vec{c}(\vec{x})=\vec{0}$ is defined as
\begin{align*}
	\vec{N}(\vec{x}) = \text{span}\{\nabla c_1(\vec{x}), \dots,\nabla c_m(\vec{x})\}
\end{align*}
and the tangent space is defined as the orthogonal complement $\vec{T}(\vec{x}) = \vec{N}(\vec{x})^\perp$. We also define the projection matrix $\vec{P_T}(\vec{x})$ that projects vectors to the tangent space $\vec{T}(\vec{x})$
\begin{align}
	\vec{P_T}(\vec{x}) = \vec{I_d }- \nabla \vec{c}(\vec{x})^T \Big(\nabla \vec{c}(\vec{x}) \nabla \vec{c}(\vec{x})^T\Big)^{-1} \nabla \vec{c}(\vec{x}) \in \mathbb{R}^{d \times d}.\label{eqn:proj-grad}
\end{align}
Then the \textit{non-degenerate index-$k$ saddle points} on the constrained manifold $\vec{c}(\vec{x}) = \vec{0}$ are special critical points of the energy functional $E(\vec{x})$ subject to the equality constraints $\vec{c}(\vec{x}) = 0$, which are defined as follows:
\begin{definition}
	Consider the case where $0<k<d-m$. A point $\vec{x^*}$ is called a non-degenerate index-$k$ saddle point of the energy functional $E(\vec{x})$ subject to $\vec{c}(\vec{x}) = \vec{0}$ if the following conditions hold:\\
	$\bullet$  1st-order KKT condition: there exists a Lagrange multiplier $\vec{\eta}(\vec{x^*}) \in \mathbb{R}^m$ with
	\begin{align}
		\nabla E(\vec{x^*}) - \vec{\eta}^T(\vec{x^*})\nabla \vec{c}(\vec{x^*}) = \vec{0};\label{eqn:KKT}
	\end{align}
	$\bullet$  second-order condition: the projected Hessian $\vec{\widehat{H}}(\vec{x^*})$ for the Lagrangian $L(\vec{x}) = E(\vec{x}) - \vec{\eta}^T(\vec{x}) \vec{c}(\vec{x})$, which is defined as
	\begin{align}
		\vec{\widehat{H}}(\vec{x^*}) = \vec{P_T}(\vec{x^*}) \Big(\nabla^2 E(\vec{x^*}) - \sum_{i=1}^m \eta_i(\vec{x^*}) \nabla^2 \vec{c}(\vec{x^*})\Big) \vec{P_T}(\vec{x^*}) \in \mathbb{R}^{d\times d}
	\end{align}
	has $k$ negative eigenvalues and $d-m-k$ positive eigenvalues in $\vec{T}(\vec{x^*})$ and the $m$-dimensional $\vec{N}(\vec{x^*})$ form the zero eigenspace of $\vec{\widehat{H}}(\vec{x^*})$.
\end{definition}
It is worth mentioning that the Lagrange multiplier $\vec{\eta}(\vec{x^*})$ in \cref{eqn:KKT} is always unique when the LICQ condition \cref{eqn:LICQ} holds at $\vec{x^*}$. The non-degeneracy of a saddle point $\vec{x^*}$ on the constrained manifold where $\vec{c}(\vec{x}) = \vec{0}$ indicates the eigenvalues on the tangent space $\vec{T}(\vec{x^*})$ never vanish. The index $k$ of the saddle point $\vec{x^*}$ indicates that there are $k$ negative directions and $m-d-k$ positive directions on the tangent space $\vec{T}(\vec{x^*})$. Applying the Morse lemma, by a change of coordinates, the energy functional $E(\vec{x})$ on the constrained manifold where $\vec{c}(\vec{x}) = \vec{0}$ can be written as a quadratic form with $k$ negative directions and $m-d-k$ positive directions. The Morse lemma (see e.g. \cite{matsumoto2002introduction}) is stated as:
\begin{lemma}[Morse lemma on constrained manifolds]\label{thm:morse-constrained}
	For a smooth energy functional $E(\vec{x}):\mathbb{R}^d \rightarrow \mathbb{R}$ and a smooth constraint function $\vec{c}(\vec{x}):\mathbb{R}^d \rightarrow \mathbb{R}^m$ on an open set $U_x \subset \mathbb{R}^d$ including $\vec{x^*}$, if the following conditions hold:\\
	$\bullet$ the LICQ condition in \cref{eqn:LICQ};\\
	$\bullet$ $\vec{x^*}$ is a non-degenerate index-$k$ saddle point for $E(\vec{x})$ on the manifold $\vec{c}(\vec{x}) = \vec{0}$.\\
	Then there exists a local coordinate $\vec{z} \in U_z \subset \mathbb{R}^{d-m}$ such that\\ $\bullet$ $\vec{x}(\vec{z}):U_z \rightarrow U_x$ satisfies $\vec{c}(\vec{x}(\vec{z}))=\vec{0}$ with $\vec{x}(0)=\vec{x^*}$;\\
	$\bullet$ the energy functional $E(\vec{x})$ with the local coordinate $z$ is in a quadratic form as $E(\vec{z}):=E(\vec{x}(\vec{z})) = E(\vec{x^*})-z_1^2-\dots-z_k^2 + z_{k+1}^2 + \dots + z_{d-m}^2$.
\end{lemma}
Using the local coordinate $\vec{z}$, we can see that $\vec{z}=\vec{0}$ is an \textit{algebraic singularity} since the level set $\mathcal{L}_{\vec{z}}= \{\vec{z} \: |\: E(\vec{z}) = E(\vec{x^*})\}$ is neither a single point nor a smooth manifold. At $\vec{z} = \vec{0}$, the Jacobian matrix of $E(\vec{z})$ is rank-deficient and not all vectors in the tangent space originate from the level set. However, in some neighborhood of $\vec{0}$ but removing the point $\vec{0}$, the level set $\mathcal{L}_{\vec{z}}$ forms a $(d-m-1)$-dimensional smooth manifold. Back to coordinate $\vec{x}$, at $\vec{x^*}$, the level set $\mathcal{L}_{\vec{x}} = \{\vec{x} \:|\: E(\vec{x}) =E(\vec{x^*})\text{ and }\vec{c}(\vec{x})=\vec{0} \}$ is neither a single point nor a smooth manifold. The saddle point $\vec{x^*}$ is an algebraic singularity and there are tangent vectors at $\vec{x^*}$ that do not originate from smooth curves within $\mathcal{L}_{\vec{x}}$.

\section{Constrained saddle search approach}\label{sec3:our-approach}
\vspace{-2ex}
Given the constrained saddle search problem in \cref{eqn:saddle-search,eqn:saddle-search-cst,eqn:pinning}, we are interested in the non-degenerate index-$k$ saddle points $\vec{p^*} \in \mathbb{R}^{nd}$ ($0<k < nd-m-d(d+1)/2$). The new bar framework corresponding to $\vec{p^*}$ has the same connection as the initial bar framework and remains under-constrained. In fact, this new bar framework is guaranteed to be singular and flexible.
\begin{theorem}\label{thm:singular-flexible}
	For the constrained saddle search problem in \cref{eqn:saddle-search,eqn:saddle-search-cst}, if $\vec{p^*}$ is a non-degenerate index-$k$ saddle point and satisfies the LICQ condition, i.e. $\nabla f_2(\vec{p^*}), \dots,\nabla f_m(\vec{p^*}), \nabla g_1(\vec{p^*}),\dots, \nabla g_{\frac{d(d+1)}{2}}(\vec{p^*})$ are linearly independent, then the bar framework corresponding to $\vec{p^*}$ is singular and flexible.
\end{theorem}
Theorem \ref{thm:singular-flexible} is a direct application of the Morse \cref{thm:morse-constrained} since the singularity in the configuration space comes from the algebraic singularity\footnote{The term \textit{algebraic singularity} here refers to a configuration where the dimension of the space of infinitesimal flexes exceeds that of the nearby constraint level sets. Readers who are interested in a detailed classification of algebraic singularity types associated with specific singular bar frameworks are referred to \cite{li2024mechanism}.} of the level set at the non-degenerate saddle point. Our goal is to utilize \cref{thm:singular-flexible} to provide a numerical algorithm to find singular and flexible bar frameworks. We provide \cref{alg:constrained-saddle-search} for searching index-$k$ saddle points in a general form with an energy functional $E(\vec{x})$ and constraints $\vec{c}(\vec{x})=\vec{0}$, which is a combination of unconstrained saddle search (see e.g. \cite{zhang2012constrained}) and Newton's method to project the search search direction back to the constraint set. However, unlike gradient descent which is guaranteed to find local minima, saddle search algorithms only have local convergence. By trying multiple initial conditions, one might be able to obtain a saddle point numerically.

\begin{algorithm}[!htb]
	\caption{Index-$k$ saddle point search of $E(\vec{x})$ with constraints $\vec{c}(\vec{x})=\vec{0}$}
	\label{alg:constrained-saddle-search}
	\begin{algorithmic}[1]
		\State \textbf{Input:} Initial values \( \vec{x_0} \), \( \vec{v_0^i} \) with $i=1,\dots,k$, step size \( \eta \), and tolerance tol.
%		\State \textbf{Output:} Sequences \( \{\vec{x_n}\} \) and \( \{\vec{v_n}\} \).
		\While{\( \|\vec{x_{n+1}}-\vec{x_n}\| > \text{tol}\)}
		\State Set $\vec{\widetilde{x}_{n+1}} = \vec{x_n} - \eta \sum_{i=1}^k (\vec{I} - 2\vec{v_n^i} \otimes \vec{v_n^i}) \vec{P_T}(\vec{x_{n}}) \nabla E(\vec{x_n})$
		
		\State Update \(\vec{ x_{n+1} }\) by projecting $\vec{\widetilde{x}_{n+1}}$ to $\vec{c}(\vec{x})=\vec{0}$ via Newton's method: $\vec{x_{n+1}} = \vec{\widetilde{x}_{n+1}} + \Big(\nabla \vec{c}(\vec{x_n})\Big)^T \vec{\alpha_n}$ with $\vec{\alpha_n}\in \mathbb{R}^m$ satisfying $\vec{c}(\vec{x_{n+1}}) = \vec{0}$.
	
		\State Update $\vec{{v}_{n+1}^i} = \vec{P_T}(\vec{x_{n+1}}) \Big(\vec{v_n^i} - (\vec{I} - \vec{v_n^i} \otimes \vec{v_n^i}- 2\sum_{j=1}^{i-1}\vec{v_n^j} \otimes \vec{v_n^j}) \vec{\widehat{H}}(\vec{x_{n+1}}) \vec{v_n^i} \Big)$ and normalize $\vec{v_{n+1}^i}$ to norm 1 with $i=1,\dots,k$.
		\EndWhile
	\end{algorithmic}
\end{algorithm}

It is worth noting that an implicit condition for the pinning scheme is needed to obtain a non-degenerate saddle point and the LICQ condition for \cref{eqn:saddle-search,eqn:saddle-search-cst,eqn:pinning}. For instance, if the pinning schemes are linearly dependent, then the LICQ condition fails automatically and the level set $f_2(\vec{p})=\dots=g_{d(d+1)/2}(\vec{p})=0$ includes some infinitesimal rigid body motions, meaning that the saddle points for the energy $f_1(\vec{p})$ are degenerate. In general, almost all pinning schemes will avoid this situation, although specifying and proving the pinning conditions required to avoid it is beyond the scope of this work, so we suggest just checking the LICQ condition and non-degeneracy of saddle points after running the saddle point search. 

\begin{remark}
	It is worth noting that numerical algebraic geometry tools, such as \textit{Bertini} (see e.g. \cite{baskar2022computing,bates2013numerically}), have been successfully used to compute saddle points in kinematic systems by solving systems of polynomial equations, particularly those with isolated zero solutions. However, Bertini's performance tends to deteriorate as the number of degrees of freedom increases. In our constrained saddle search problem, the number of variables is not small due to the combination of free vertices and Lagrange multipliers associated with the constraints, which makes Bertini less practical for this setting.
\end{remark}

\textbf{Examples of singular and flexible bar frameworks:} We present two examples of 2D singular and flexible bar frameworks identified through our constrained saddle search approach in \cref{fig:heptagon-1}. Let us explain in detail using the example in \cref{fig:heptagon-1}(a)-(e). We begin with a non-singular heptagon consisting of 10 edges, where non-trivial nonlinear flexes forms a one-dimensional smooth manifold. To avoid rigid body motion in 2D, here we fix three coordinates, the $x, y$ coordinates of vertex $A$ and the $y$ coordinate of vertex $G$. Next, we set the edge $BF$ free and this additional degree of freedom allows us to search for index-1 saddle points of the energy functional, the squared length of edge $BF$. By searching for index-1 saddle point numerically, we obtain a bar framework in \cref{fig:heptagon-1}(b), which does not have the four-bar linkage as a subgraph and has a two-dimensional non-trivial infinitesimal flexes. By testing the non-degeneracy and the LICQ condition, we find that the new bar framework is singular and only two infinitesimal flexes (up to scaling) that come from nonlinear flexes. These two infinitesimal flexes are found by a second-order stress test and plotted in \cref{fig:heptagon-1}(b) and (d). We also find the nonlinear flexes corresponding to these two infinitesimal flexes, and plot one deformed states along the nonlinear flexes in \cref{fig:heptagon-1}(c) and (e)\footnote{Animations of these nonlinear flexes and more examples of singular and flexible frameworks are available at \url{https://xuenanli.github.io/research/}.}.

\begin{figure}[!htb]
	\centering
	\subfloat[]{
		\begin{tikzpicture}[scale=0.35]
			\pgfmathsetmacro{\xscale}{1.1} % Define the x-axis scaling constant
			
			% Define the vertices of the final state
			\coordinate (A2) at ({\xscale*0},0);
			\coordinate (B2) at ({\xscale*(-1.2)}, 1.34);
			\coordinate (C2) at ({\xscale*(-1.11)}, 2.63);
			\coordinate (D2) at ({\xscale*0.39}, 5.23);
			\coordinate (E2) at ({\xscale*2.56}, 3.98);
			\coordinate (F2) at ({\xscale*3.62}, 2.92);
			\coordinate (G2) at ({\xscale*1}, 0);

			% Draw the boundary of the heptagon
			\draw[ultra thick] (A2) -- (B2);
			\draw[ultra thick] (B2) -- (C2);
			\draw[ultra thick] (C2) -- (D2);
			\draw[ultra thick] (D2) -- (E2);
			\draw[ultra thick] (E2) -- (F2);
			\draw[ultra thick] (F2) -- (G2);
			\draw[ultra thick] (G2) -- (A2);
			
			% connect the vertices by 3 interior edges 
			\draw[ultra thick, red, dashed] (B2) -- (F2);
			\draw[ultra thick] (A2) -- (C2);
			\draw[ultra thick] (D2) -- (G2);

			% Label the points
			\node[below left] at (A2) {A};
			\node[above left] at (B2) {B};
			\node[above left] at (C2) {C};
			\node[above left] at (D2) {D};
			\node[above] at (E2) {E};
			\node[above right] at (F2) {F};
			\node[below] at (G2) {G};
			
			% Draw vertices for the heptagon - LAST instruction
			\fill[black] (A2) circle (5pt);
			\fill[black] (G2) circle (5pt);
			\fill[black] (B2) circle (5pt);
			\fill[black] (C2) circle (5pt);
			\fill[black] (D2) circle (5pt);
			\fill[black] (E2) circle (5pt);
			\fill[black] (F2) circle (5pt);
		\end{tikzpicture}
	}
	\subfloat[]{
		\begin{tikzpicture}[scale=0.4]
			\pgfmathsetmacro{\xscale}{1.1} % Define the x-axis scaling constant
			
			% Define the vertices of the final state
			\coordinate (A2) at ({\xscale*0},0);
			\coordinate (B2) at ({\xscale*0.776635405826052}, 1.62383418070239);
			\coordinate (C2) at ({\xscale*2.01192626375719}, 2.02887306317142);
			\coordinate (D2) at ({\xscale*3.35090912329313}, 4.71348195495975);
			\coordinate (E2) at ({\xscale*1.61101477589929}, 2.91827509782407);
			\coordinate (F2) at ({\xscale*0.465391046208262}, 3.88654482779027);
			\coordinate (G2) at ({\xscale*1}, 0);
			
			% Define the flex vectors
			\coordinate (A3) at ({\xscale*0},0);
			\coordinate (B3) at ({\xscale*0.776635405826052}, 1.62383418070239);
			\coordinate (C3) at ({\xscale*2.01192626375719}, 2.02887306317142);
			\coordinate (D3) at ({\xscale*(3.35090912329313-0.6)}, 4.71348195495975+0.3);
			\coordinate (E3) at ({\xscale*(1.61101477589929-0.4)}, 2.91827509782407+0.15);
			\coordinate (F3) at ({\xscale*(0.465391046208262-0.6)}, 3.88654482779027-0.1);
			\coordinate (G3) at ({\xscale*1}, 0);
			
			\def\k{1.5} % Define the vector scaling constant k
			
			% Define the flexes scaled by k
			\coordinate (A4) at ($(A2) + \k*(A3) - \k*(A2)$);
			\coordinate (B4) at ($(B2) + \k*(B3) - \k*(B2)$);
			\coordinate (C4) at ($(C2) + \k*(C3) - \k*(C2)$);
			\coordinate (D4) at ($(D2) + \k*(D3) - \k*(D2)$);
			\coordinate (E4) at ($(E2) + \k*(E3) - \k*(E2)$);
			\coordinate (F4) at ($(F2) + \k*(F3) - \k*(F2)$);
			\coordinate (G4) at ($(G2) + \k*(G3) - \k*(G2)$);
			
			% Draw the boundary of the heptagon
			\draw[ultra thick] (A2) -- (B2);
			\draw[ultra thick] (B2) -- (C2);
			\draw[ultra thick] (C2) -- (D2);
			\draw[ultra thick] (D2) -- (E2);
			\draw[ultra thick] (E2) -- (F2);
			\draw[ultra thick] (F2) -- (G2);
			\draw[ultra thick] (G2) -- (A2);
			
			% connect the vertices by 3 interior edges 
			\draw[ultra thick] (B2) -- (F2);
			\draw[ultra thick] (A2) -- (C2);
			\draw[ultra thick] (D2) -- (G2);
			
			% make vectors
			\draw[-{Latex[length=2mm, width=2mm]},thick, red] (D2) -- (D4);
			\draw[-{Latex[length=2mm, width=2mm]},thick, red] (E2) -- (E4);
			\draw[-{Latex[length=2mm, width=2mm]},thick, red] (F2) -- (F4);
			
			% Label the points
			\node[below] at (A2) {A};
			\node[above left] at (B2) {B};
			\node[right] at (C2) {C};
			\node[above] at (D2) {D};
			\node[above] at (E2) {E};
			\node[above] at (F2) {F};
			\node[below] at (G2) {G};
			
			% Draw vertices for the heptagon - LAST instruction
			\fill[black] (A2) circle (5pt);
			\fill[black] (G2) circle (5pt);
			\fill[black] (B2) circle (5pt);
			\fill[black] (C2) circle (5pt);
			\fill[black] (D2) circle (5pt);
			\fill[black] (E2) circle (5pt);
			\fill[black] (F2) circle (5pt);
			
		\end{tikzpicture}
	}
	\subfloat[]{
		\begin{tikzpicture}[scale=0.4]
			\pgfmathsetmacro{\xscale}{1.1} % Define the x-axis scaling constant
			
			% New, less clustered framework coordinates
			\coordinate (A2) at ({\xscale*0},0);
			\coordinate (B2) at ({\xscale*0.733937858235272}, 1.64357391688023);
			\coordinate (C2) at ({\xscale*1.95822284364036}, 2.08075382779981);
			\coordinate (D2) at ({\xscale*3.70650466279132}, 4.51868546769996);
			\coordinate (E2) at ({\xscale*1.35787262610640}, 3.66198974793871);
			\coordinate (F2) at ({\xscale*-0.139298491153350}, 3.75406930942322);
			\coordinate (G2) at ({\xscale*1}, 0);

			% Draw the boundary of the heptagon
			\draw[ultra thick] (A2) -- (B2);
			\draw[ultra thick] (B2) -- (C2);
			\draw[ultra thick] (C2) -- (D2);
			\draw[ultra thick] (D2) -- (E2);
			\draw[ultra thick] (E2) -- (F2);
			\draw[ultra thick] (F2) -- (G2);
			\draw[ultra thick] (G2) -- (A2);
			
			% connect the vertices by 3 interior edges 
			\draw[ultra thick] (B2) -- (F2);
			\draw[ultra thick] (A2) -- (C2);
			\draw[ultra thick] (D2) -- (G2);
			
			% Label the points
			\node[below] at (A2) {A};
			\node[shift={(0.1cm, 0.35cm)}] at (B2) {B};
			\node[above] at (C2) {C};
			\node[above] at (D2) {D};
			\node[above] at (E2) {E};
			\node[above] at (F2) {F};
			\node[below] at (G2) {G};
			
			% Draw vertices for the heptagon - LAST instruction
			\fill[black] (A2) circle (5pt);
			\fill[black] (G2) circle (5pt);
			\fill[black] (B2) circle (5pt);
			\fill[black] (C2) circle (5pt);
			\fill[black] (D2) circle (5pt);
			\fill[black] (E2) circle (5pt);
			\fill[black] (F2) circle (5pt);
		\end{tikzpicture}
	}
	\subfloat[]{
		\begin{tikzpicture}[scale=0.4]
			\pgfmathsetmacro{\xscale}{1.1} % Define the x-axis scaling constant
			
			% Define the vertices of the final state
			\coordinate (A2) at ({\xscale*0},0);
			\coordinate (B2) at ({\xscale*0.776635405826052}, 1.62383418070239);
			\coordinate (C2) at ({\xscale*2.01192626375719}, 2.02887306317142);
			\coordinate (D2) at ({\xscale*3.35090912329313}, 4.71348195495975);
			\coordinate (E2) at ({\xscale*1.61101477589929}, 2.91827509782407);
			\coordinate (F2) at ({\xscale*0.465391046208262}, 3.88654482779027);
			\coordinate (G2) at ({\xscale*1}, 0);
			
			% Define the flex vectors
			\coordinate (A3) at ({\xscale*0.136224815426610}, -0.321100649655604);
			\coordinate (B3) at ({\xscale*0.358931478032394}, 1.56766248297513);
			\coordinate (C3) at ({\xscale*1.45605388804108}, 2.39408864027066);
			\coordinate (D3) at ({\xscale*(3.41366388407085-0.8)}, 4.77014937026241+0.3);
			\coordinate (E3) at ({\xscale*(1.85607453621453+0.15)}, 2.79825455266347-0.4);
			\coordinate (F3) at ({\xscale*(0.858703186676733+0.6)}, 3.94193144390966+0.1);
			\coordinate (G3) at ({\xscale*1.13622482652173}, 0.0200232840221762);
			
			\def\k{1} % Define the vector scaling constant k
			
			% Define the flexes scaled by k
			\coordinate (A4) at ($(A2) + \k*(A3) - \k*(A2)$);
			\coordinate (B4) at ($(B2) + \k*(B3) - \k*(B2)$);
			\coordinate (C4) at ($(C2) + \k*(C3) - \k*(C2)$);
			\coordinate (D4) at ($(D2) + \k*(D3) - \k*(D2)$);
			\coordinate (E4) at ($(E2) + \k*(E3) - \k*(E2)$);
			\coordinate (F4) at ($(F2) + \k*(F3) - \k*(F2)$);
			\coordinate (G4) at ($(G2) + \k*(G3) - \k*(G2)$);
			
			% Draw the boundary of the heptagon
			\draw[ultra thick] (A2) -- (B2);
			\draw[ultra thick] (B2) -- (C2);
			\draw[ultra thick] (C2) -- (D2);
			\draw[ultra thick] (D2) -- (E2);
			\draw[ultra thick] (E2) -- (F2);
			\draw[ultra thick] (F2) -- (G2);
			\draw[ultra thick] (G2) -- (A2);
			
			% connect the vertices by 3 interior edges 
			\draw[ultra thick] (B2) -- (F2);
			\draw[ultra thick] (A2) -- (C2);
			\draw[ultra thick] (D2) -- (G2);
			
			% Label the points
			\node[below left] at (A2) {A};
			\node[above left] at (B2) {B};
			\node[right] at (C2) {C};
			\node[above left] at (D2) {D};
			\node[above] at (E2) {E};
			\node[above] at (F2) {F};
			\node[below] at (G2) {G};
			
			% Draw vertices for the heptagon - LAST instruction
			\fill[black] (A2) circle (5pt);
			\fill[black] (G2) circle (5pt);
			\fill[black] (B2) circle (5pt);
			\fill[black] (C2) circle (5pt);
			\fill[black] (D2) circle (5pt);
			\fill[black] (E2) circle (5pt);
			\fill[black] (F2) circle (5pt);
			
			% make vectors
			\draw[-{Latex[length=2mm, width=2mm]},thick, red] (D2) -- (D4);
			\draw[-{Latex[length=2mm, width=2mm]},thick, red] (E2) -- (E4);
			\draw[-{Latex[length=2mm, width=2mm]},thick, red] (F2) -- (F4);
		\end{tikzpicture}
	}
	\subfloat[]{
		\begin{tikzpicture}[scale=0.4]
			\pgfmathsetmacro{\xscale}{1.1} % Define the x-axis scaling constant
			
			% Define the vertices of the final stat
			\coordinate (A2) at ({\xscale*0},0);
			\coordinate (B2) at ({\xscale*0.705195767209454}, 1.65610957666148);
			%0.109024394788571, 1.79669521103079);
			\coordinate (C2) at ({\xscale*1.92167939840661}, 2.11454994904309);
			%1.10134671962304, 2.63651447946604);
			\coordinate (D2) at ({\xscale*3.75443992330031}, 4.48962654957440);
			%3.18930253716010, 4.79068264932027);
			\coordinate (E2) at ({\xscale*1.94115051542503}, 2.76858572400189);
			%1.81541577872930, 2.70203831027916);
			\coordinate (F2) at ({\xscale*0.983506769631487}, 3.92310660131096);
			%0.943696701856928, 3.92273722934839);
			\coordinate (G2) at ({\xscale*1}, 0);
			
			% Draw the boundary of the heptagon
			\draw[ultra thick] (A2) -- (B2);
			\draw[ultra thick] (B2) -- (C2);
			\draw[ultra thick] (C2) -- (D2);
			\draw[ultra thick] (D2) -- (E2);
			\draw[ultra thick] (E2) -- (F2);
			\draw[ultra thick] (F2) -- (G2);
			\draw[ultra thick] (G2) -- (A2);
			
			% connect the vertices by 3 interior edges 
			\draw[ultra thick] (B2) -- (F2);
			\draw[ultra thick] (A2) -- (C2);
			\draw[ultra thick] (D2) -- (G2);
			
			% Label the points
			\node[below] at (A2) {A};
			\node[left] at (B2) {B};
			\node[shift={(-0.15cm,0.15cm)}] at (C2) {C};
			\node[above] at (D2) {D};
			\node[shift={(0cm,0.35cm)}] at (E2) {E};
			\node[above] at (F2) {F};
			\node[below] at (G2) {G};
			
			% Draw vertices for the heptagon - LAST instruction
			\fill[black] (A2) circle (5pt);
			\fill[black] (G2) circle (5pt);
			\fill[black] (B2) circle (5pt);
			\fill[black] (C2) circle (5pt);
			\fill[black] (D2) circle (5pt);
			\fill[black] (E2) circle (5pt);
			\fill[black] (F2) circle (5pt);
		\end{tikzpicture}
	}\\
	\subfloat[]{
		\begin{tikzpicture}[scale=0.45]
			% Define the vertices of the final state
			\coordinate (A2) at (0,0);
			\coordinate (B2) at (-0.0291611389336360, 1.79976377004764);
			\coordinate (C2) at (-0.674484071819545, 2.92828417849913);
			\coordinate (D2) at (1.15521103685379, 2.12068538621620);
			\coordinate (E2) at (2.75549472728296, 0.921063733299569);
			\coordinate (F2) at (3.70391276750481, 1.23808616307659);
			\coordinate (G2) at (1, 0);
			
			% Define the flex vectors
			\coordinate (A3) at (0,0);
			\coordinate (B3) at (-0.0291611389336360, 1.79976377004764);
			\coordinate (C3) at (-0.674484071819545+0.42, 2.92828417849913+0.24);
			\coordinate (D3) at (1.15521103685379+0.8, 2.12068538621620+1);
			\coordinate (E3) at (2.75549472728296-0.12, 0.921063733299569+0.3);
			\coordinate (F3) at (3.70391276750481, 1.23808616307659);
			\coordinate (G3) at (1, 0);
			
			\def\k{2} % Define the vector scaling constant k
			
			% Define the flexes scaled by k
			\coordinate (A4) at ($(A2) + \k*(A3) - \k*(A2)$);
			\coordinate (B4) at ($(B2) + \k*(B3) - \k*(B2)$);
			\coordinate (C4) at ($(C2) + \k*(C3) - \k*(C2)$);
			\coordinate (D4) at ($(D2) + \k*(D3) - \k*(D2)$);
			\coordinate (E4) at ($(E2) + \k*(E3) - \k*(E2)$);
			\coordinate (F4) at ($(F2) + \k*(F3) - \k*(F2)$);
			\coordinate (G4) at ($(G2) + \k*(G3) - \k*(G2)$);
			
			% Draw the boundary of the heptagon
			\draw[ultra thick] (A2) -- (B2);
			\draw[ultra thick] (B2) -- (C2);
			\draw[ultra thick] (C2) -- (D2);
			\draw[ultra thick] (D2) -- (E2);
			\draw[ultra thick] (E2) -- (F2);
			\draw[ultra thick] (F2) -- (G2);
			\draw[ultra thick] (G2) -- (A2);
			
			% connect the vertices by 3 interior edges 
			\draw[ultra thick] (B2) -- (F2);
			\draw[ultra thick] (A2) -- (E2);
			\draw[ultra thick] (C2) -- (G2);
			
			% Label the points
			\node[below right] at (A2) {A};
			\node[shift={(-0.3cm,0.05cm)}] at (B2) {B};
			\node[above] at (C2) {C};
			\node[shift={(-0.1cm,0.3cm)}] at (D2) {D};
			\node[shift={(-0.5cm,0.1cm)}] at (E2) {E};
			\node[right] at (F2) {F};
			\node[below right] at (G2) {G};
			
			% Draw vertices for the heptagon - LAST instruction
			\fill[black] (A2) circle (4pt);
			\fill[black] (G2) circle (4pt);
			\fill[black] (B2) circle (4pt);
			\fill[black] (C2) circle (4pt);
			\fill[black] (D2) circle (4pt);
			\fill[black] (E2) circle (4pt);
			\fill[black] (F2) circle (4pt);
			
			% make vectors
			\draw[-{Latex[length=2mm, width=2mm]},thick, red] (C2) -- (C4); 
			\draw[-{Latex[length=2mm, width=2mm]},thick, red] (D2) -- (D4);
			\draw[-{Latex[length=2mm, width=2mm]},thick, red] (E2) -- (E4);
		\end{tikzpicture}
	}
	\subfloat[]{
		\begin{tikzpicture}[scale=0.45]
			% Define the vertices of the final state
			\coordinate (A2) at (0,0);
			\coordinate (B2) at (-0.183477173692298, 1.79062450746489);
			%-0.360497988711100, 1.76353089004286);
			\coordinate (C2) at (-0.388026772765720, 3.07443113679772);
			%-0.414339712463571, 3.06241543794672);
			\coordinate (D2) at (1.61183892383709, 3.05125365739669);
			%1.55616570894984, 3.40462646083005);
			\coordinate (E2) at (2.59380885818236, 1.30891756810680);
			%2.42347098708198, 1.60246621420114);
			\coordinate (F2) at (3.57889482780143, 1.48098044167245);
			%3.41449509522947, 1.73614948566947);
			\coordinate (G2) at (1, 0);
			
			% Draw the boundary of the heptagon
			\draw[ultra thick] (A2) -- (B2);
			\draw[ultra thick] (B2) -- (C2);
			\draw[ultra thick] (C2) -- (D2);
			\draw[ultra thick] (D2) -- (E2);
			\draw[ultra thick] (E2) -- (F2);
			\draw[ultra thick] (F2) -- (G2);
			\draw[ultra thick] (G2) -- (A2);
			
			% connect the vertices by 3 interior edges 
			\draw[ultra thick] (B2) -- (F2);
			\draw[ultra thick] (A2) -- (E2);
			\draw[ultra thick] (C2) -- (G2);
			
			% Label the points
			\node[below] at (A2) {A};
			\node[left] at (B2) {B};
			\node[above] at (C2) {C};
			\node[above] at (D2) {D};
			\node[shift={(-0.5cm, 0cm)}] at (E2) {E};
			\node[above] at (F2) {F};
			\node[below] at (G2) {G};
			
			% Draw vertices for the heptagon - LAST instruction
			\fill[black] (A2) circle (4pt);
			\fill[black] (G2) circle (4pt);
			\fill[black] (B2) circle (4pt);
			\fill[black] (C2) circle (4pt);
			\fill[black] (D2) circle (4pt);
			\fill[black] (E2) circle (4pt);
			\fill[black] (F2) circle (4pt);
		\end{tikzpicture}
	}
	\subfloat[]{
		\begin{tikzpicture}[scale=0.45]
			% Define the vertices of the final state
			\coordinate (A2) at (0,0);
			\coordinate (B2) at (-0.0291611389336360, 1.79976377004764);
			\coordinate (C2) at (-0.674484071819545, 2.92828417849913);
			\coordinate (D2) at (1.15521103685379, 2.12068538621620);
			\coordinate (E2) at (2.75549472728296, 0.921063733299569);
			\coordinate (F2) at (3.70391276750481, 1.23808616307659);
			\coordinate (G2) at (1, 0);
			
			% Define the flex vectors
			\coordinate (A3) at (0,0);
			\coordinate (B3) at (-0.0291611389336360, 1.79976377004764);
			\coordinate (C3) at (-0.674484071819545+1, 2.92828417849913+0.6);
			\coordinate (D3) at (1.15521103685379+1.2, 2.12068538621620+1);
			\coordinate (E3) at (2.75549472728296+0.2, 0.921063733299569-0.4);
			\coordinate (F3) at (3.70391276750481, 1.23808616307659);
			\coordinate (G3) at (1, 0);
			
			\def\k{1.5} % Define the vector scaling constant k
			
			% Define the flexes scaled by k
			\coordinate (A4) at ($(A2) + \k*(A3) - \k*(A2)$);
			\coordinate (B4) at ($(B2) + \k*(B3) - \k*(B2)$);
			\coordinate (C4) at ($(C2) + \k*(C3) - \k*(C2)$);
			\coordinate (D4) at ($(D2) + \k*(D3) - \k*(D2)$);
			\coordinate (E4) at ($(E2) + \k*(E3) - \k*(E2)$);
			\coordinate (F4) at ($(F2) + \k*(F3) - \k*(F2)$);
			\coordinate (G4) at ($(G2) + \k*(G3) - \k*(G2)$);
			
			% Draw the boundary of the heptagon
			\draw[ultra thick] (A2) -- (B2);
			\draw[ultra thick] (B2) -- (C2);
			\draw[ultra thick] (C2) -- (D2);
			\draw[ultra thick] (D2) -- (E2);
			\draw[ultra thick] (E2) -- (F2);
			\draw[ultra thick] (F2) -- (G2);
			\draw[ultra thick] (G2) -- (A2);
			
			% connect the vertices by 3 interior edges 
			\draw[ultra thick] (B2) -- (F2);
			\draw[ultra thick] (A2) -- (E2);
			\draw[ultra thick] (C2) -- (G2);
			
			% Label the points
			\node[below right] at (A2) {A};
			\node[shift={(-0.3cm,0.05cm)}] at (B2) {B};
			\node[above] at (C2) {C};
			\node[shift={(-0.05cm,0.3cm)}] at (D2) {D};
			\node[shift={(-0.5cm,0.1cm)}] at (E2) {E};
			\node[right] at (F2) {F};
			\node[below right] at (G2) {G};
			
			% Draw vertices for the heptagon - LAST instruction
			\fill[black] (A2) circle (4pt);
			\fill[black] (G2) circle (4pt);
			\fill[black] (B2) circle (4pt);
			\fill[black] (C2) circle (4pt);
			\fill[black] (D2) circle (4pt);
			\fill[black] (E2) circle (4pt);
			\fill[black] (F2) circle (4pt);
			
			% make vectors
			\draw[-{Latex[length=2mm, width=2mm]},thick, red] (C2) -- (C4); 
			\draw[-{Latex[length=2mm, width=2mm]},thick, red] (D2) -- (D4);
			\draw[-{Latex[length=2mm, width=2mm]},thick, red] (E2) -- (E4);
		\end{tikzpicture}
	}
	\subfloat[]{
		\begin{tikzpicture}[scale=0.45]
			% Define the vertices of the final state
			\coordinate (A2) at (0,0);
			\coordinate (B2) at (-0.469993623972845, 1.73755747917157);
			\coordinate (C2) at (-0.466973418647420, 3.03755397084851);
			\coordinate (D2) at (1.50376101316838, 2.69666425369132);
			\coordinate (E2) at (2.69352907948906, 1.08904117958473);
			\coordinate (F2) at (3.30232016853687, 1.88237176004798);
			\coordinate (G2) at (1, 0);
			
			% Draw the boundary of the heptagon
			\draw[ultra thick] (A2) -- (B2);
			\draw[ultra thick] (B2) -- (C2);
			\draw[ultra thick] (C2) -- (D2);
			\draw[ultra thick] (D2) -- (E2);
			\draw[ultra thick] (E2) -- (F2);
			\draw[ultra thick] (F2) -- (G2);
			\draw[ultra thick] (G2) -- (A2);
			
			% connect the vertices by 3 interior edges 
			\draw[ultra thick] (B2) -- (F2);
			\draw[ultra thick] (A2) -- (E2);
			\draw[ultra thick] (C2) -- (G2);
			
			% Label the points
			\node[below] at (A2) {A};
			\node[left] at (B2) {B};
			\node[above] at (C2) {C};
			\node[above] at (D2) {D};
			\node[below right] at (E2) {E};
			\node[above right] at (F2) {F};
			\node[below] at (G2) {G};
			
			% Draw vertices for the heptagon - LAST instruction
			\fill[black] (A2) circle (4pt);
			\fill[black] (G2) circle (4pt);
			\fill[black] (B2) circle (4pt);
			\fill[black] (C2) circle (4pt);
			\fill[black] (D2) circle (4pt);
			\fill[black] (E2) circle (4pt);
			\fill[black] (F2) circle (4pt);
		\end{tikzpicture}
	}
	\caption{(a) initial non-singular heptagon; (b)(d) final state with two special infinitesimal flexes; (c)(e) deformed states along the two directions in (b)(d). Another singular and flexible bar framework is shown in figures (f)-(i), following the same instructions as those for figures (b)-(e).}
	\label{fig:heptagon-1}
\end{figure}

\vspace{-4ex}
\section{Conclusions and future work}\label{sec4:conclusion}
\vspace{-2ex}
In this paper, we interpret singularities in flexible bar frameworks as non-\linebreak degenerate saddle points on constrained manifolds and provide a numerical algorithm to obtain new singular and flexible frameworks. Future directions include: (1) exploring 3D singular bar frameworks; (2) allowing multiple edges to vary their lengths to uncover singular frameworks with higher-index; (3) generalizing our approach to periodic frameworks for identifying singular lattice systems and designing novel metamaterials (see e.g. \cite{li2023some,li2025effective,li2025nonlinear}); and (4) generalizing it to broader geometric constraints to explore singular structures in complex robotic systems.

%\clearpage
\bibliographystyle{spmpsci}
\bibliography{ref}% common bib file

\end{document}